\documentclass[letterpaper, 10 pt, conference]{ieeeconf}  
\renewcommand{\baselinestretch}{0.95}

\IEEEoverridecommandlockouts                              

\usepackage{graphicx} 
\usepackage{amsmath} 
\usepackage{url}
\usepackage{tikz} 
\UseRawInputEncoding

\title{\LARGE \bf
EMG-based Control Strategies of a Supernumerary Robotic Hand\\
for the Rehabilitation of Sub-Acute Stroke Patients: Proof of Concept
}

\author{Marina Gnocco\authorrefmark{4}$^{1}$, Manuel G. Catalano $^{1}$, Giorgio Grioli $^{1,4}$, Carlo Trompetto $^{2,3}$, Antonio Bicchi$^{1,4}$
\thanks{*This work was supported by the European Research Council Synergy Grant Natural BionicS (NBS) project (Grant Agreement No. 810346)}
\thanks{$^{1}$Soft Robotics for Human Cooperation and Rehabilitation lab, Fondazione Istituto Italiano di
Tecnologia, Genova 16163, Italy}%
\thanks{$^{2}$ Department of Neuroscience, Rehabilitation, Ophthalmology, Genetics, Maternal and Child Health,
University of Genova, Genova 16132, Italy.}
\thanks{$^{3}$Neurorehabilitation Unit, Department of Neuroscience, IRCCS Ospedale Policlinico San Martino,
Genova, Genova 16132, Italy.}
\thanks{$^{4}$Centro di Ricerca “Enrico Piaggio” and Dipartimento di Ingegneria dell’Informazione, Università di
Pisa, Pisa 56122, Italy}
 \thanks{\authorrefmark{4} Corresponding author; mail: \tt\small marina.gnocco@iit.it}
}

\newcommand\copyrighttext{%
  \footnotesize \textcopyright\ \the\year{} IEEE. Personal use of this material is permitted.  Permission from IEEE must be obtained for all other uses, in any current or future media, including reprinting/republishing this material for advertising or promotional purposes, creating new collective works, for resale or redistribution to servers or lists, or reuse of any copyrighted component of this work in other works.}

\newcommand\copyrightnotice{%
\begin{tikzpicture}[remember picture,overlay]
\node[anchor=south,yshift=10pt] at (current page.south) {\fbox{\parbox{\dimexpr\textwidth-\fboxsep-\fboxrule\relax}{\copyrighttext}}};
\end{tikzpicture}%
}

\begin{document}

\maketitle
\thispagestyle{empty}
\pagestyle{empty}

\begin{abstract}
One of the most frequent and severe aftermaths of a stroke is the loss of upper limb functionality. 
Therapy started in the sub-acute phase proved more effective, mainly when the patient participates actively.  
Recently, a novel set of rehabilitation and support
robotic devices, known as supernumerary robotic limbs, have been introduced.
This work investigates how a surface electromyography (sEMG) based control strategy would improve their usability in rehabilitation, limited so far by input interfaces requiring to subjects some level of residual mobility.

 After briefly introducing the phenomena hindering post-stroke sEMG and its use to control robotic hands, we describe a framework to acquire and interpret muscle signals of the forearm extensors. We applied it to drive a supernumerary robotic limb, the SoftHand-X, to provide Task-Specific Training (TST) in patients with sub-acute stroke. We propose and describe two algorithms to control the opening and closing of the robotic hand, with different levels of user agency and therapist control. 
We experimentally tested the feasibility of the proposed approach on four patients, followed by a therapist, to check their ability to operate the hand. 
The promising preliminary results indicate sEMG-based control as a viable solution to extend TST to sub-acute post-stroke patients.
\end{abstract}

\copyrightnotice

\section{Introduction}
Upper limb dysfunction represents a common complication after a stroke. 
It affects 80\% of subjects early after stroke \cite{Nakayama} and results in permanent disability between 23\% \cite{Nakayama} and 50\% \cite{Kwakkel2003} of times.
Recovery, when possible, occurs because the brain reorganizes itself after the injury so that the healthy tissue takes over the compromised functions of the damaged area. 
This phenomenon, known as "neuroplasticity", is most beneficial during the initial three months after the stroke (referred to as the sub-acute phase). Extensive evidence \cite{Langhorne2011, Zeiler2013} has shown that commencing rehabilitation during this phase leads to superior outcomes in long-term motor recovery. Active patient participation in the training of the impaired limb is a necessary prerequisite in this process. 

%

\begin{figure}[t!]
      \centering
      \includegraphics[width=1\columnwidth]{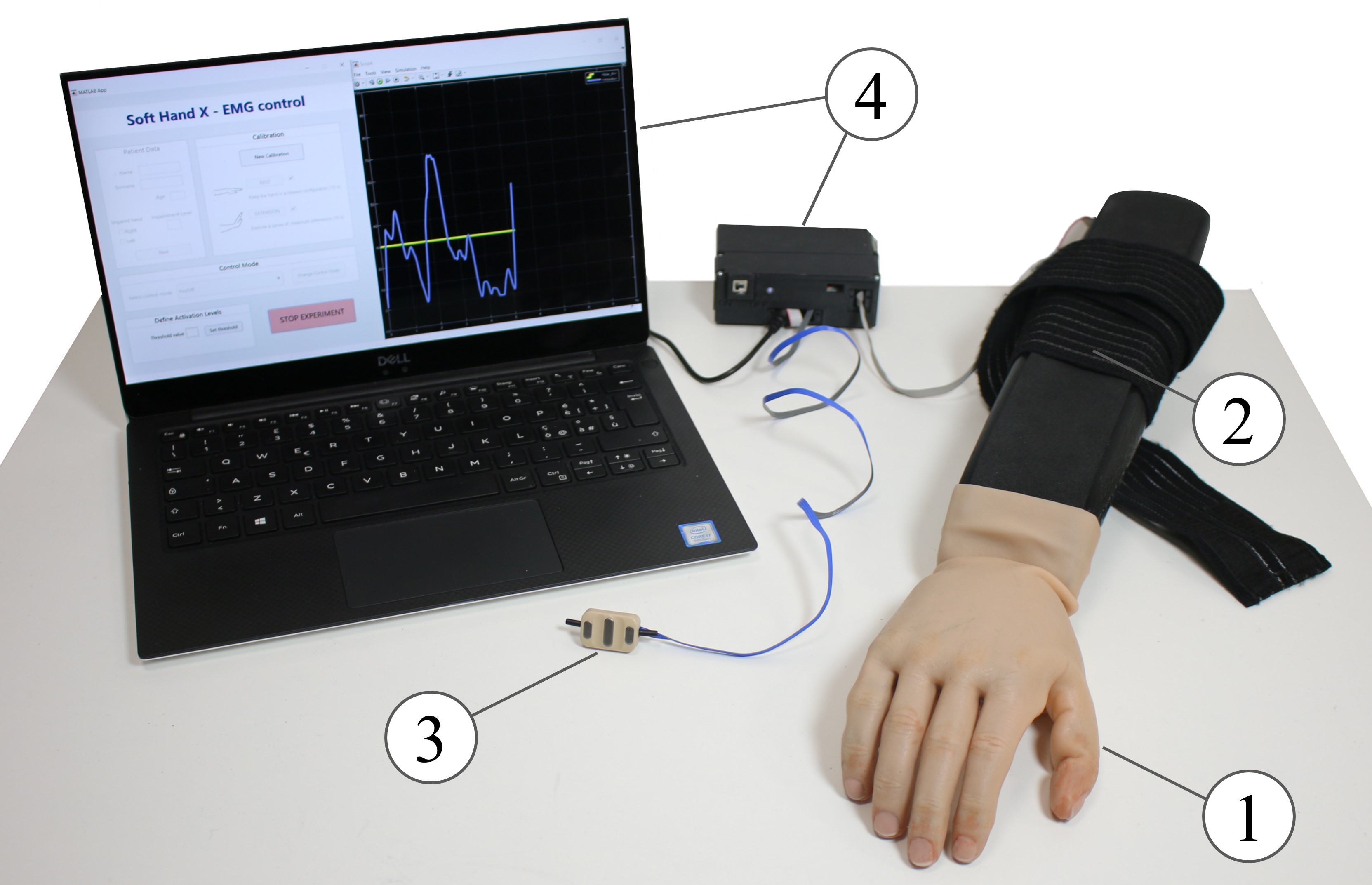}
      \caption{Experimental and training hardware setup of the EMG-driven SoftHand-X system. The figure shows the sub-parts constituting the overall architecture: the robotic hand (1), the human-arm interface (2), the sEMG electrode (3), and the remote workstation (4), including laptop, battery, and the controller for analog acquisition, data exchange, and motor control.}
      \label{Fig2}
   \end{figure}

Post-stroke Task-Specific Training (TST), proved very effective for the functional recovery of the upper limb \cite{Good2011}. It draws inspiration from the principle of motor learning, and it is based on intensive goal-directed motor tasks that the patient has to accomplish repeatedly using the impaired limb (e.g., reaching, grasping, etc.).
At least two factors, however, hamper TST in the sub-acute phase. First, most patients do not present sufficient muscular force to perform meaningful goal-directed motor tasks in this recovery phase. For these patients, hand rehabilitation is usually limited to passive mobilization and mirror therapy \cite{Gandhi2020}.
Second, the repetitive and intense flexion movements of fingers and wrist typical of TST are thought to favor the onset of spasticity \cite{Laidler1994}, a pathologic condition characterized by an exaggeration of the stretch reflex. Spasticity, mainly affecting flexor muscles, can lead to a long-term disability if not treated properly. 

Robotic devices have been demonstrated as an effective tool in rehabilitating sub-acute stroke patients. However, typical interfaces either use the non-affected limb to control the device or require some residual movement of the impaired limb to work \cite{maciejasz}. Therefore, their use is limited to subjects with a certain grade of residual mobility, excluding a large slice of potential users among post-stroke survivors. This work investigates how a surface electromyography (sEMG) based control strategy would improve the usability of robotic tools in rehabilitation. 

In a previous work \cite{Trompetto123}, we provided evidence of the practicability of a supernumerary robotic limb, the SoftHand-eXtrathesis (SoftHand-X), as a rehabilitative tool to provide safe task-specific therapy for sub-acute stroke patients. Thanks to its control strategy, the SoftHand-X allows performing TST without requiring the activation of flexor muscles, thus avoiding spasticity enhancement. 

Moreover, the aspect and movements of the SoftHand-X are inspired by the human hand. During the therapy with the SoftHand-X, patients reported having had the illusion that the hand they were employing was their real one \cite{Trompetto123}. This illusion is supposed to stimulate the same brain areas activated during well-established therapy for stroke rehabilitation like mirror therapy \cite{Gandhi2020} and Action Observation Therapy (AOT) \cite{Buccino2014}, i.e., the mirror neuron system.

This work aims to extend the usability of the SoftHand-X to subjects in which the mobility is reduced or completely absent using sEMG-based control strategies. We propose using sEMG to identify action intention from the muscle activity of the patients, which often persists in the absence of visible motion, even for the most severe impairments of the sub-acute phase. 

The paper is organized as follows: Sec. II presents the strengths and issues of using EMG-controlled devices in post-stroke rehabilitation. Sec. III describes the hardware setup employed for the study and, in Sec. IV, the proposed control algorithms are presented. Sec. V provides the results of experiments conducted on four patients during TST, discussed in Sec. VI. Finally, conclusions are drawn in Sec. VII.

\section{Post-Stroke sEMG Signal Recovery}
After a stroke, EMG can detect electrical muscle activity even before any movement is observable.
Therefore, it represents a viable alternative to identify motion intention when limb mobility is severely compromised, to replace assistive devices relying on the detection of kinematic signals. 
However, soon after the episode, patients tend to develop abnormal muscle activation in the form of muscle weakness, spastic hypertonia, and impaired movement coordination, which contaminate voluntary EMG activity \cite{Dewald2001}. 
Consequently, in existing literature, the emphasis of EMG-driven devices for the upper limb lies primarily in addressing the chronic phase of stroke rehabilitation, wherein muscle control is typically enhanced. \cite{maciejasz}.
In those studies, multi-channel EMG is generally used to detect contractions of different muscles. 
Pattern recognition algorithms are then applied to discriminate between different classes of motions on the model of prosthetic control.
While these strategies accurately recognize motion intention within amputees, contradictory results have emerged about their effectiveness with subjects with neurological injuries, with recognition accuracy rated between 25\% \cite{Cesqui2013} and 65\% \cite{ramos}.
Moreover, better recognition accuracy was obtained when EMG data were processed with powerful, time-consuming algorithms \cite{Zhang2012, Yu}.
For real-time applications, models recognizing a limited set of gestures are best suited since they require fewer input channels and limit the computational cost.
Usually, such models also feature more intuitive threshold-based control strategies. 

SoftHand-X characteristics are in line with these specifications. Designed only to allow one degree of actuation, a single input is sufficient for its control. Therefore, it is
reasonable to consider it a good platform for developing an EMG-based control strategy.
Our model employs two primary control approaches for EMG signals, on-off and proportional, which have been extensively studied and documented in the literature.
In the on-off control mode, a function of the device is turned on or off (e.g., either constant speed in one direction, full stop, or constant speed in the other direction) \cite{Fougner}. The robustness and intuitiveness of this control strategy explain its continuing popularity \cite{Fougner}. Proportional control allows instead the user to continuously control a mechanical output quantity of the actuator (e.g., force, velocity, position, etc.) by varying the control input within a corresponding continuous interval. It allows positioning the terminal device much more precisely than possible with on-off control \cite{Fougner}.
 Due to the non-idealities of the pathologic EMG signal following stroke, some corrections have been proposed in our model to ensure a robust control.

\section{The SoftHand-X System}
The supernumerary system presented in this work is composed of four functional subsystems (see Fig. \ref{Fig2}): (1) A robotic hand, (2) a human-arm interface to fasten the robotic hand to the patient's forearm securely, (3) sEMG electrodes, and (4) a remote workstation to control and monitor the system. 
The software and the electronic framework behind the system architecture are derived from the open-source platform Natural Machine Motion Initiative \cite{Santina2011}\footnote{NMMI website platform: www.naturalmachinemotioninitiative.com}.

\subsection{Robotic hand}
The end-effector employed derives from the Pisa/IIT SoftHand Pro (SHP) \cite{Godfrey2018}, originally developed for prosthetic applications. 
It is an anthropomorphic robotic hand with 19 degrees of freedom.
A single motor moves all the hand joints according to the principle of under-actuation, which replicates the synergistic behavior of the human hand. A single input is, therefore, enough for its control.
The soft structure of the hand allows the finger to autonomously adapt its movement to the objects to be grasped or manipulated, providing a secure grip despite different shapes or sizes. 
In this way, there is no need to select different hand grips for different tasks, and the cognitive effort requested to the user to control the hand is minimized. All these features make the hand an excellent candidate to be used by subjects with neurological injuries.
The hand integrates the DC motor driving its motion and the custom electronic board needed for its control.
\subsection{Human-arm interface} 
The robotic hand is secured to the patient's arm through a rigid splint fixed tightly to the robotic hand. 
The splint is 3D-printed in Acrylonitrile Butadiene Styrene (ABS). 
The whole system can be fastened to the patient's forearm through a wide elastic band that allows a firm hold while reducing slippage. 
The design of this interface was optimized, with respect to the version employed in \cite{Ciullo2020}, removing the bulky gravity compensator in favor of the therapist's support, and to the one used in \cite{Trompetto123}, to distribute the weight of the robotic hand evenly and fit the forearm more comfortably.

\subsection{sEMG electrodes}
The acquisition setup is based on Ottobock 13-E200, double differential surface EMG electrodes commonly used in prosthetics.
These sensors provide an adjustable signal amplification (2.000-100.000), on board rectification, band-pass filtering in the range of 90 Hz-450 Hz, and power line interference suppression. 
Output is an analog signal between 0-5V already enveloped.

\subsection{Workstation}
To minimize the load on the user, the battery powering the system is installed in a decentralized module. 
sEMG electrodes are also connected to this module, which integrates the microcontroller (PSoC 5, Cypress), managing analog signal acquisition and serial data exchange between the robotic hand and PC.
The controller transmits and receives data to and from the PC by a serial connection.
On the remote PC, a Matlab/Simulink program performs input processing and generates the control signal for the motor. 
A Graphical User Interface (GUI) has been implemented to simplify these operations. 
The therapists can use the GUI during the therapy to monitor the patient's EMG signal and adjust threshold levels regulating hand activation.
The GUI also allows setting up the proper calibration, recording the sEMG signal during the entire session, and starting/stopping the system. 

\section{Proposed Approach}
SoftHand-X is programmed to control its aperture through the extension of the wrist and fingers, while closure happens automatically when the muscles are relaxed.
This allows to safe administer TST in the sub-acute phase, as the recruitment of flexor muscles is avoided, which is a cause of spasticity enhancement.
We propose a very simple control system for the SoftHand-X, but effective, based on a single sEMG sensor, placed on the impaired upper limb in correspondence with wrist and fingers extensor muscles (extensor carpi radialis, extensor carpi ulnaris).
To complete TST, the use of a single electrode is, in fact, enough to detect the patient's intention to move while bypassing issues related to abnormal muscular activation patterns in post-stroke patients. 
As the patient is only asked to extend the wrist and fingers to drive the SoftHand-X, an increase in the EMG amplitude recorded from extensor muscles can only correspond to a voluntary extension intent.
This would not be true if two antagonist muscles, and two electrodes, were employed to drive the opening and closing of the SoftHand-X.
In such a case, because of co-contraction issues, an increased signal detected by the electrode placed on the extensors could have also been caused by the voluntary contraction of flexors and vice-versa.   

\begin{figure}[hbt]
      \centering
    \includegraphics[width=1\columnwidth]{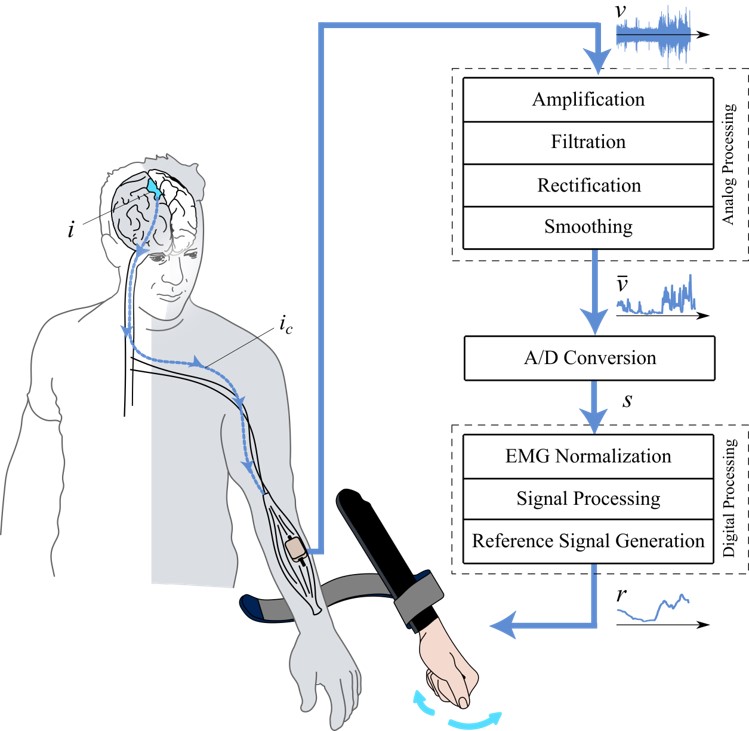}
      \caption{Schematic of signal processing steps involved in the control of the SoftHand-X. As soon as a motor intent (\(i\)) is triggered in the stroke patient's brain, a corrupted neural signal (\(i_C\)) is delivered to the muscle involved, causing a weak contraction. sEMG electrodes can transduce this activity in a measurable electrical potential (\(v\)). After a conditioning phase, in which it is filtered and rectified (\(\bar v\)), the signal is sampled by an Analog to Digital Converter (\(s\)) and fed to the PC to be processed. The output is a control signal (\(r\)) sent to the motor to drive the robotic hand. }
      \label{Fig3}
   \end{figure}

A newly developed computer-based control platform (in Fig. \ref{Fig3}) allowed the simultaneous real-time detection of the sEMG levels and position control of the SoftHand-X. 
As soon as the user expresses a motor intention (\(i\)), a corrupted neural signal (\(i_C\)) is delivered from the brain to the muscle involved, causing a weak contraction that sEMG can identify. 
Once recorded, the sEMG signal needs to be processed.
Raw sEMG signal (\(v\)) underwent a conditioning phase of amplification and filtering. 
Chosen Ottobock electrodes already integrate these steps and provide a rectified, enveloped analog output between 0-5 V, with frequency components in the range of 90-450 Hz (\(\bar v\)).
The PSoC sampled the signal at 1Khz with a 12-bit resolution.
The digital signal was then transmitted to the PC.

Data are processed within the Simulink environment, and the robotic hand's reference position (\(r\)) is generated. 
A magnetic encoder reads the actual position reached by the motor and closes the control loop. 
A calibration phase is required before applying any EMG-based control strategy to the hand. 
EMG signals during maximum voluntary contraction (MVC) and rest phase are measured and used to normalize the input. 
Finally, a moving average on 50 samples is applied to smooth the signal.
At each discrete time \(k\), we get a signal \(EMG k\), to which the control algorithms are applied.

At rest, the robotic hand is fully closed. 
Two strategies have been proposed to control the finger opening:
\subsection{On-off}
This type of control causes the SoftHand-X to open entirely as soon as \(EMG_k\) exceeds a given threshold \(th\). 
The hand keeps this position as long as the signal stays above the threshold; otherwise, it closes again. 
We chose not to fix the activation threshold to a predetermined value but to leave the clinician to set it according to the patient's signal variability. 
The user interface allows the therapist intuitively to tune the threshold to EMG signal intensity, which is displayed in real-time.

The control signal obtained is the following:
\begin{equation}
 r_k=
    \begin{cases}
      1, & \text{if}\ EMG_k>th \\
      0, & \text{otherwise}
    \end{cases}      
\end{equation}
where \(r_k\) is the reference position sent to the motor at time \(k\) normalized between \(0\), when the hand is fully closed, and \(1\) when it is fully open.
On-off control is particularly promising for this study since it requires the patient to make minimal efforts to keep the SoftHand-X fully opened. 
It is, therefore, suitable also for subjects with more severe motor deficits.

\subsection{sEMG-Proportional}
In the proportional control, the SoftHand-X is programmed to vary its opening level proportionally to the intensity of the recorded EMG signal. 
The proportionality is guaranteed in a range of EMG intensity identified between two levels:
\begin{itemize}
    \item A lower threshold sets the minimum muscular signal useful to activate the SoftHand-X. It is needed to remove uncertainty at low contraction levels \(th_1\);
    \item A higher threshold identifies the EMG intensity needed to provoke a full aperture of the SoftHand-X. In this way, SoftHand-X can be fully opened using a level of contraction lower than the maximum the patient can perform, reducing fatigue \(th_2\).
\end{itemize}
Once again, threshold levels are not fixed but can be selected by the therapist as required on a case-by-case basis.

The resulting reference control signal is:
\begin{gather}
r_k=p \cdot EMG_k\\
p=\frac{EMG_k-th_1}{th_2-th_1}
\end{gather}
Proportional control is applicable if the user can precisely control their muscular contraction continuously, which may not be trivial after a stroke. 
Fluctuations of the EMG signal may result in unwanted ripples in the SoftHand-X aperture level, thus compromising the natural aspect of the robotic hand.
Here, a correction on the signal is applied to neutralize the impact of unwanted EMG fluctuations on the control of the hand and stabilize its motion. 

\begin{figure}[htb]
      \centering
      \includegraphics[width=0.8\columnwidth]{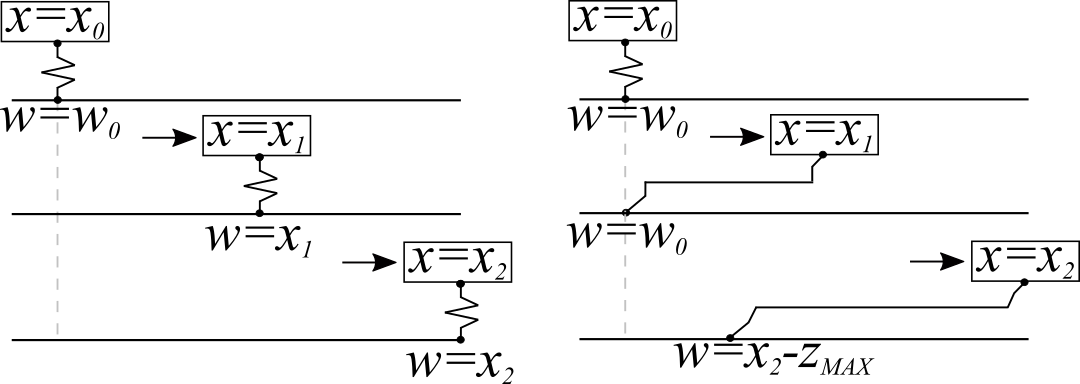}
      \caption{Conceptual explanation of Dahl's model for friction \cite{Hayward}. When friction is not considered, any force horizontally acting on a moving object placed on a fixed surface will make it slip (left). Introducing friction forces in the system, adhesion may occur between the object and the surface, causing a change in the object's motion (right). Dahl's model is exploited in this study to operate a correction on proportional control.}
      \label{Fig5}
   \end{figure}

To define terms for modeling this correction strategy, consider the mechanical analogy presented in \cite{Hayward}, shown in Fig. \ref{Fig5}.
A moving object has a point of contact with a fixed surface.
Dahl models friction by considering two points, one belonging to the moving object (\(x\)) and one defining its contact point to the surface, called \(w\).
In the absence of friction, any horizontal force applied to the moving object will make it slip on the surface together with the contact point \(w\), so that \(z=x-w=0\) at each time \(k\). 
If friction is introduced in the system, during adhesion \(w\) is attached to the surface, and the application of a horizontal force will provoke relative movement between the two points (\(z=x-w>0\)). 
Friction can be modeled as a spring connecting the two points, and friction force as proportional to strain \(z\). 
At \(z=zmax\), the contact becomes fully tense, and \(w\) relocates so that at all times \(|z|=zmax\) and the object slides.

We apply the same concept to our problem, with a small but significant change of  perspective. 
Indeed, we look at the motions of the point attached to the fixed surface as a regularization of the motions of the object. 
By stating
\begin{gather}
   p \cdot EMG=x\\
  r=w \\
  \Delta=z_{max}
\end{gather}
where \(\Delta\) is the maximum fluctuation intensity of the EMG signal, considered an unwanted ripple and not a voluntary contraction variation.
We obtain the following reference control signal for the hand:
\begin{equation}
 r_{k+1}=
    \begin{cases}
      x_k+\Delta, & \text{if}\ x_k-r_k<-\Delta \\
      r_k, & \text{if}\ |x_k-r_k|\leq\Delta \\
      x_k-\Delta, & \text{if}\ x_k-r_k>\Delta \\
    \end{cases}      
\end{equation}

It is then re-scaled through multiplication to adapt the signal span to the hand aperture.
\begin{equation}
 r_k=\frac{r_k-\Delta}{100-2\Delta} \cdot 100    
\end{equation}
In this case, all the input fluctuations of intensity lower than \(\Delta\) have no impact on the resulting reference signal at the cost of a slight latency in the system response.
Therefore, this solution is more robust against EMG non-idealities, and its use is promising with stroke patients.

\section{Experimental Validation}

For this explorative study, the efficacy and usability of the two control strategies were tested with four sub-acute stroke patients (two males and two females, aged between 60 and 75). 
Selected subjects had a stroke no longer than two months before the enrolment and presented minimal or no ability to extend the fingers of the impaired hand.  
Patients who did not present detectable voluntary EMG signals on the muscles of interest were excluded from the study. 
Patients recruited were able to understand and follow the instructions provided during the experiment and gave written informed consent to the study.
Patients were positioned in a seated posture facing a table. The right or left version of the SoftHand-X was employed according to the patient's impaired side and fastened to the forearm using the elastic band as described in Sec.III.b. 
The skin was cleaned and shaved to optimize recording, and the electrode positioned in correspondence with the main activity spot of the wrist and finger, identified with a therapist's help.
Surgical tape was used to keep the electrode in position. 
Electrode analog gain was set to the maximum (100.000) because of the patients' EMG signal weakness. %
\begin{figure}[t!]
      \centering
    \includegraphics[width=\columnwidth]{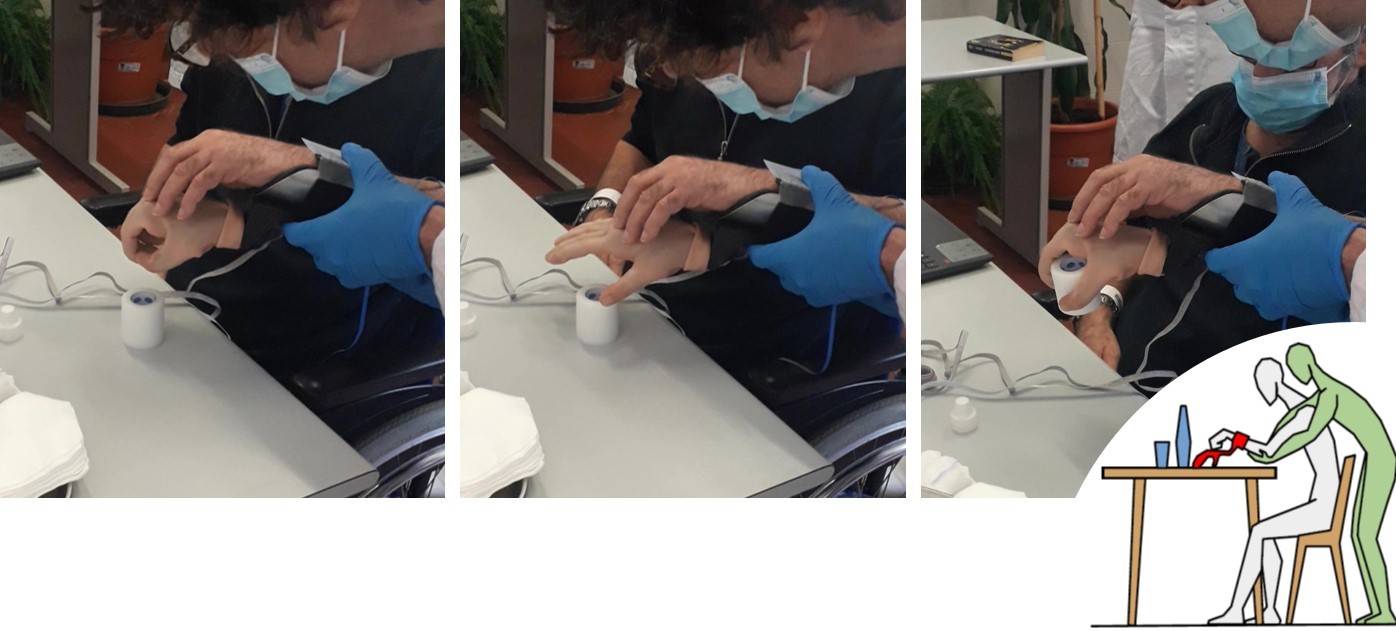}
      \caption{Stroke patient executing grasping tasks using the EMG-controlled SoftHand-X during the experimental phase.
      The physiotherapist supports the arm and forearm, helping the patient in the proximal movements.}
          \label{Fig7}
   \end{figure}
In the initial calibration phase, patients were firstly asked to keep the arm relaxed, to record the minimum EMG level, and then to perform a series of maximal extensions of the selected muscles to record the maximum EMG signal. This procedure was needed to normalize the EMG signal during the experiment and to set an initial activation threshold at half the EMG dynamic range, specific for each patient and each session.

During the experiment, the patient had to modulate wrist and finger extension to control the SoftHand-X and complete the grasping actions. Each session had a duration of approximately 20 minutes. Objects employed included foam balls, wooden cubes normally used for Box and Blocks test \cite{konston}, cylindrical shapes, and surgical tape.
The physiotherapist sustained the device's weight and helped the patient maintain a good posture during movements, following the protocol explained in \cite{Trompetto123}. The photo sequence in Fig. \ref{Fig7} clarifies the experiment's setup.
During the experimental session, on-off control and corrected proportional control were tested.

\section{Discussion}
Despite preliminary, trial results proved sEMG to be a viable solution to identify motion intentions even in severely impaired stroke patients. 
In this regard, the simple recording setup based on a single electrode demonstrated to be robust against stroke-related EMG corruption and functional to sub-acute TST. 
Moreover, thanks to the therapist's help, the robot hand's weight did not impact the quality of the EMG recording during the training. 
Eventually, the use of a single electrode detecting, in general, muscle potentials of various extensor muscles in the forearm did not require the electrode positioning to be very precise to ensure good detection accuracy, thus proving another point in favor of this recording setup.
The post-stroke motor recovery, moreover, starts at the proximal level, and with rehabilitation, it proceeds towards the most distal areas of the body.
In this way, even more severe patients, able to activate the extensor muscles of the wrist, but not yet those of the fingers, can be recruited for the TST with the SoftHand-x in the sub-acute phase.
   \begin{figure}[t]
      \centering
    \includegraphics[width=1\columnwidth]{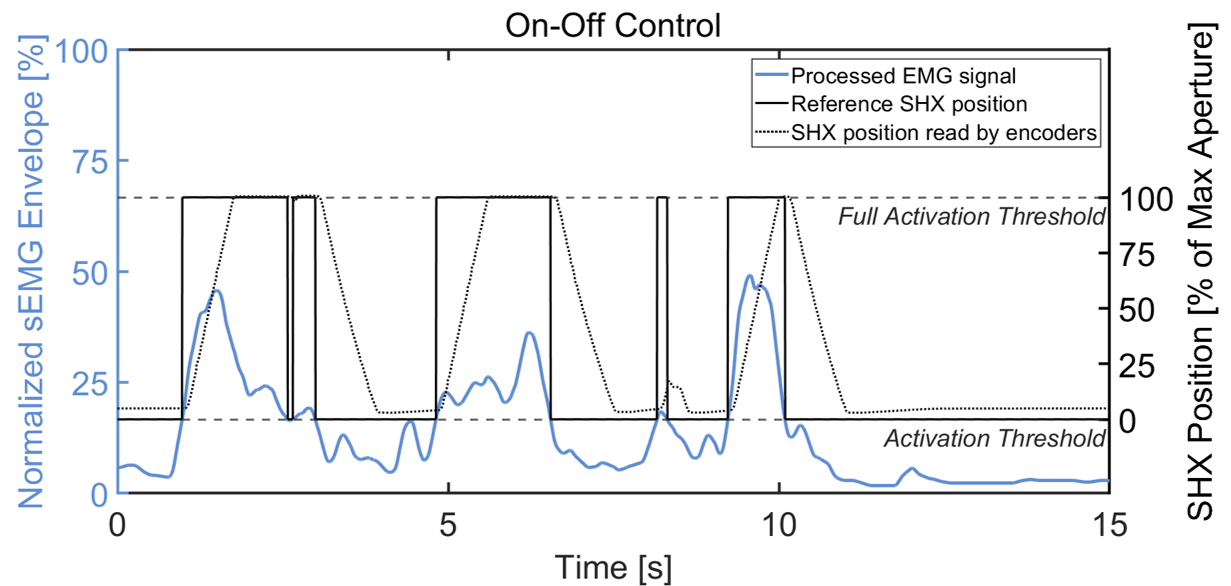}\vspace{10pt}
        \includegraphics[width=1\columnwidth]{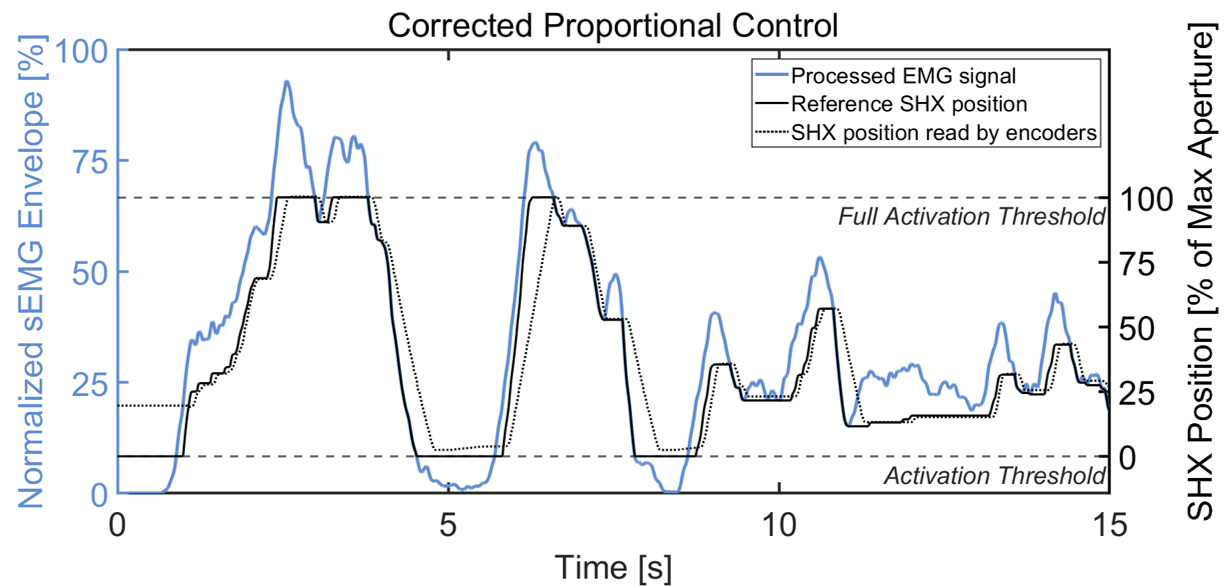}
        \caption{Data acquired during the experimental tests with the on-off (top) and the corrected proportional (bottom) control strategies.
In each plot, the left vertical axis is associated with the conditioned EMG
input (cyan line), horizontal dashed lines indicate the activation thresholds and
the right vertical axis to the position of the SoftHand-X as a percentage of the maximum aperture. The continuous black line indicates the control signal sent to the motor with the given input, while the dotted black line shows the actual position reached by the SoftHand-X, as reported by the encoders.
The horizontal axis indicates times.}
      \label{Fig8}
   \end{figure}
When the on-off control strategy was employed, all the patients enrolled could efficiently complete simple grasping/releasing tasks. 
Indeed, as it only allows two states of the hand, this control did not require a perfectly constant level of activation to keep the hand position. 
Keeping the EMG level above the activation threshold is enough to get the hand stably open. 
Because of this, even one subject classified with MRC grade 1 (contraction without visible movement of wrist and finger) was able to perform the exercise successfully.

However, the low effort required is also the main concern raised by on-off control: the patient, even if able to perform stronger contractions, may get used to performing the minimum helpful contraction to cross the threshold, as the extra effort would not correspond to further hand aperture, thus reducing the effectiveness of rehabilitation.
On the contrary, the corrected proportional control was supposed to provide more interaction during the training, as it reproduces a natural hand's entire range of motion. 
However, the proportional control resulted more challenging for the patients, even with the applied correction. 
The performance  largely depended on the user's ability to stabilize the contraction level, disadvantaging the ones affected by more critical motor impairments. 
Fig. \ref{Fig8} shows the EMG recording of a patient with a mild reduction of hand functionality. 
He managed to exploit proportional control, adjusting muscle contraction but, despite the correction, he was unable to keep the robotic hand stable enough to hold the object after it had been grasped. 
This type of control may be used in the later stages of rehabilitation to sharpen fine hand movements when one would expect a partial recovery of muscle strength. 
On-off control resulted instead the most appropriate for the sub-acute phase. 
Future development would also consider making the control of the correction parameter \(\Delta\) available to the therapist so that the filtering power on the EMG signal could be customized on the patient.

Moreover, other advantages emerged in favor of the on-off control strategy. 
Therapists considered it effective since it required minimum effort, and therapy sessions could be prolonged. 
This is particularly useful for TST, which requires repeating a single task as many times as possible to promote neural plasticity. 
Moreover, an excessive effort during extension, because of the inability of the pathological subjects to independently recruit antagonist muscles, might lead to simultaneous, stronger activation of the flexor muscles. 
Promoting flexors activation is not recommended in case of an observable onset of hypertonus. 
Control approaches requiring a lower level of muscle activation are therefore preferable. 
As shown in Fig. \ref{Fig8} (top), also the on-off strategy is not immune to unwanted movements of the SoftHand-X. 
Oscillations of the EMG activity around the threshold may provoke jerk movements of the robotic hand. 
In the future, we are planning to implement a hysteresis on the activation threshold to solve this issue.

Finally, the medical staff particularly appreciated the possibility of customizing the device's activation threshold.
Contraction abilities may differ significantly from patient to patient, and a fixed threshold at a given percentage of MVC could have been unsuitable for all of them to trigger the hand. 
Therapists found it helpful to change the threshold in real-time without needing a new calibration during the same exercise session. 
In fact, the EMG signal intensity reduced during the session because of fatigue. 
The adjustable threshold prevented this from causing the exercise to stop. The therapist’s evaluation showed no sign of increased spasticity after the treatment.

Comments on the patients’ use experience were collected. They appreciated the training with the SoftHand-X in general and reported a good embodiment of the robotic hand. They did not find the training difficult to execute nor expressed fatigue. 

\section{Conclusions}
In this work, we investigated the use of the patient’s residual muscular activity, recorded with sEMG, to drive a supernumerary robotic limb, the SoftHand-X, for robot-assisted sub-acute stroke TST. 
We proposed and described two algorithms to control the opening and closing of the robotic hand, with different levels of user agency and therapist control.
Feedback from the therapist was encouraging, and the patient demonstrated the capability to control and operate the SoftHand-X.
Better results have been obtained with one of the two proposed control strategies, the on-off one. 
The positive outcomes of this study suggest the possibility of performing sEMG-driven task-specific hand training already from the sub-acute phase using supernumerary limbs.
The main limitation of the study was the reduced sample of subjects involved in the experimental test, and an extended clinical trial is programmed to validate these preliminary results quantitatively.



\section*{Acknowledgement}
The authors would like to thank Vinicio Tincani, Mattia Poggiani, Cristiano Petrocelli and Manuel Barbarossa for their help in the design and realization of the SoftHand-X hardware.
Thanks to Luca Puce for his support in the experimental activities.

\bibliographystyle{IEEEtran}
\bibliography{IEEEabrv,mybibfile}

\end{document}